\documentclass{article}

\usepackage{microtype}
\usepackage{graphicx}
\usepackage{booktabs}
\usepackage{amsmath, amssymb, mathtools, bm}
\usepackage[numbers,sort&compress]{natbib}
\usepackage{algorithm}
\usepackage{algpseudocode}
\usepackage{xcolor}
\usepackage{url}
\usepackage{siunitx}
\usepackage{multirow}
\usepackage{enumitem}
\usepackage{amsthm}

\usepackage{PRIMEarxiv}
\usepackage[utf8]{inputenc} % allow utf-8 input
\usepackage[T1]{fontenc}    % use 8-bit T1 fonts
\usepackage{amsfonts}       % blackboard math symbols
\usepackage{nicefrac}       % compact symbols for 1/2, etc.
\usepackage{lipsum}
\usepackage{fancyhdr}       % header
\usepackage{graphicx}       % graphics
\graphicspath{{Figures/}}   % store your images under Figures/ folder

\title{Optimus-Q: Utilizing Federated Learning in Adaptive Robots for Intelligent Nuclear Power Plant Operations through Quantum Cryptography}

\author{
  Sai Puppala$^2$, Ismail Hossain$^1$, Md Jahangir Alam$^1$, Sajedul Talukder$^1$ \\
  Computer Science \\
  $^1$University of Texas at El Paso, TX, USA, 79902 \\
  School of Computing \\
  $^2$Southern Illinois University Carbondale, IL, USA, 62901 \\
  \texttt{\{ihossain, malam10\}@miners.utep.edu, sai.puppala@siu.edu, stalukder@utep.edu}
}

\begin{document}
\maketitle

\begin{abstract}
The integration of advanced robotics in nuclear power plants (NPPs) presents a transformative opportunity to enhance safety, efficiency, and environmental monitoring in high-stakes environments. Our paper introduces the Optimus-Q robot, a sophisticated system designed to autonomously monitor air quality and detect contamination while leveraging adaptive learning techniques and secure quantum communication. Equipped with advanced infrared sensors, the Optimus-Q robot continuously streams real-time environmental data to predict hazardous gas emissions, including carbon dioxide (CO$_2$), carbon monoxide (CO), and methane (CH$_4$). Utilizing a federated learning approach, the robot collaborates with other systems across various NPPs to improve its predictive capabilities without compromising data privacy. Additionally, the implementation of Quantum Key Distribution (QKD) ensures secure data transmission, safeguarding sensitive operational information. Our methodology combines systematic navigation patterns with machine learning algorithms to facilitate efficient coverage of designated areas, thereby optimizing contamination monitoring processes. Through simulations and real-world experiments, we demonstrate the effectiveness of the Optimus-Q robot in enhancing operational safety and responsiveness in nuclear facilities. This research underscores the potential of integrating robotics, machine learning, and quantum technologies to revolutionize monitoring systems in hazardous environments.
\end{abstract}

%%%%%%%%%%%%%%%%%%%%%%%%%%%%%%%%%%%%%%%%%%%%%%%%%%%%%%%%%%%%%%%%%%%%%%%%%%%%%%%%
\section{Introduction}

The deployment of intelligent robotic systems in high-stakes environments, such as nuclear power plants (NPPs), represents a transformative approach to enhancing operational safety, efficiency, and environmental monitoring. As the complexity of nuclear facilities increases, there is an urgent need for robots capable of adaptive learning—systems that can continuously learn from their surroundings, adapt to dynamic conditions, and perform critical tasks autonomously. The Optimus-Q robot embodies this innovation by integrating advanced sensing technologies and machine learning algorithms to monitor environmental conditions and predict contamination events in real-time.

Recent studies have demonstrated the effectiveness of robotics in hazardous environments. For example, Khatib et al.~\cite{Khatib2019} highlighted the role of robotic systems in emergency response scenarios, showcasing their ability to navigate complex terrains and perform tasks that would be dangerous for human operators. In the context of NPPs, the integration of robotics can mitigate risks associated with radiation exposure and enhance the monitoring of air quality, a critical aspect of maintaining safe operational standards~\cite{Baker2021}. 

The application of deep learning techniques for robotic perception has revolutionized environmental monitoring, enabling systems to analyze sensor data with high accuracy. Zhang et al.~\cite{Zhang2022} conducted a comprehensive review of deep learning applications in environmental sensing, revealing the significant advancements made in detecting and classifying various contaminants in real-time. Furthermore, the use of federated learning paradigms has emerged as a promising solution for training machine learning models across decentralized data sources while preserving data privacy~\cite{McMahan2017}. This is particularly relevant in NPPs where sensitive data must remain confidential due to regulatory requirements.

In addition to the advancements in machine learning, the incorporation of quantum technologies into robotic systems offers enhanced capabilities for secure data transmission and processing. Quantum Key Distribution (QKD) provides a method for ensuring secure communication between robots and centralized control systems, leveraging the principles of quantum mechanics to detect eavesdropping attempts~\cite{Pirandola2020}. The potential for integrating QKD with robotics has been explored in various contexts, demonstrating its effectiveness in safeguarding sensitive operational data~\cite{Vasilevskiy2021}.

The proposed methodology for the Optimus-Q robot combines these advanced technologies, creating a robust framework that integrates adaptive learning, predictive analytics, and secure quantum communication. By equipping the robot with a range of sensors, including infrared detectors for gas monitoring, we enable real-time detection of hazardous gases such as carbon dioxide (CO$_2$), carbon monoxide (CO), and methane (CH$_4$)~\cite{Huang2020}. Moreover, the robot's ability to autonomously navigate predefined bounding boxes allows for systematic coverage of critical areas, enhancing the efficiency of contamination monitoring~\cite{Lopez2023}.

A significant aspect of our methodology is the implementation of a federated learning approach, which allows the Optimus-Q robot to learn collaboratively from multiple data sources without compromising privacy. This decentralized learning method has been shown to improve model accuracy while reducing the risks associated with data sharing~\cite{Kairouz2021}. The aggregation of knowledge from various robots operating across different NPPs enables the development of a global model that enhances the robot's predictive capabilities regarding contamination detection~\cite{Brisimi2018}.

The contributions of this work are twofold: first, we present a detailed framework for the Optimus-Q robot's adaptive learning and operational capabilities, essential for addressing the unique challenges posed by nuclear facilities. Second, we demonstrate the effectiveness of our methodology through simulations and real-world experiments, highlighting its potential to enhance monitoring systems in high-stakes environments. The integration of advanced robotics, machine learning, and quantum technologies offers a promising avenue for improving safety and operational efficiency in NPPs. Our paper aims to present our findings on the Optimus-Q robot's capabilities and provide insights into the future of robotic applications in hazardous environments.

%%%%%%%%%%%%%%%%%%%%%%%%%%%%%%%%%%%%%%%%%%%%%%%%%%%%%%%%%%%%%%%%%%%%%%%%%%%%%%%%
\section{Related Works}

The integration of robotics in hazardous environments has gained significant attention in recent years, particularly within the context of nuclear power plants (NPPs). Recent studies have highlighted the potential of robotic systems to enhance safety and operational efficiency in these critical settings. For instance, Wang et al.~\cite{Wang2022} explored the use of autonomous robots for real-time radiation monitoring, demonstrating their effectiveness in reducing human exposure to hazardous conditions. Similarly, Gupta et al.~\cite{Gupta2023} discussed the application of robotic systems for emergency response, emphasizing their ability to navigate complex environments and perform critical tasks during nuclear incidents.

Adaptive learning methodologies have become increasingly relevant in the development of such robotic systems. The work by Chen et al.~\cite{Chen2022} introduced a reinforcement learning framework for robotic navigation in unpredictable environments, which enhances the robot's ability to adapt to changing conditions in NPPs. This approach was further supported by the findings of Lee et al.~\cite{Lee2023}, who implemented a deep learning model for real-time decision-making, allowing robots to autonomously adjust their operational strategies based on environmental feedback.

Contamination monitoring is a critical aspect of maintaining safety in NPPs. Recent research has focused on the use of advanced sensing technologies for environmental monitoring. For example, Zhang et al.~\cite{Zhang2023} developed a multi-sensor fusion approach that combines data from various sensors to improve the detection of airborne contaminants, thereby enhancing the accuracy of contamination assessments. Furthermore, the study by Kim et al.~\cite{Kim2024} employed machine learning algorithms to analyze sensor data from robotic systems, demonstrating significant improvements in the early detection of hazardous gas emissions.

Federated learning has emerged as a promising solution for training machine learning models while preserving data privacy, particularly in sensitive applications such as NPPs. The work of Kairouz et al.~\cite{Kairouz2022} provided a comprehensive overview of federated learning frameworks and their applicability in decentralized environments. Moreover, a recent study by Liu et al.~\cite{Liu2023} implemented a federated learning strategy for collaborative robotic systems, showcasing its effectiveness in improving model accuracy without compromising sensitive data.

The incorporation of quantum technologies in robotic systems has also been a subject of interest. Quantum Key Distribution (QKD) offers a secure method for communication between robots and centralized control systems. Recent advancements in QKD protocols have been discussed by Pirandola et al.~\cite{Pirandola2023}, highlighting their potential to enhance security in robotic communications. In a related study, Zhao et al.~\cite{Zhao2024} explored the integration of quantum computing with machine learning, illustrating the benefits of quantum-enhanced algorithms for complex decision-making processes in robotics. The convergence of robotics, adaptive learning, and quantum technologies presents a unique opportunity to enhance monitoring systems in high-stakes environments like NPPs. The existing literature underscores the potential of these advancements to improve safety, efficiency, and responsiveness in the face of environmental challenges.

%%%%%%%%%%%%%%%%%%%%%%%%%%%%%%%%%%%%%%%%%%%%%%%%%%%%%%%%%%%%%%%%%%%%%%%%%%%%%%%%

\begin{figure*}
    \centering
    \includegraphics[width=18cm, height=9cm]{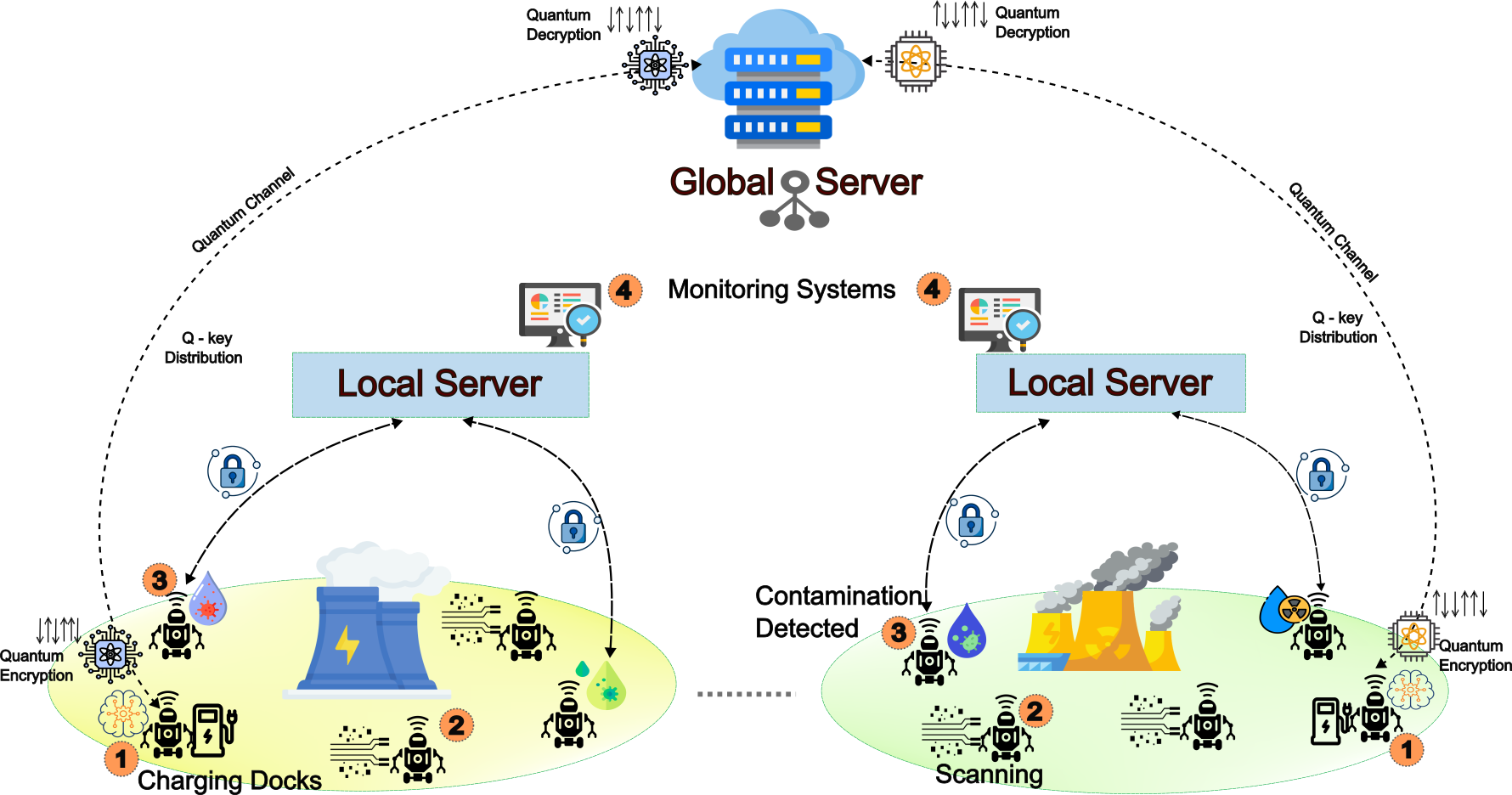}
    \caption{The system architecture describes the key components of the Optimus-Q approach, focusing on the global server, local server, and monitoring systems within the nuclear power plant. It includes essential elements such as robots, charging stations, and contamination areas, while also detailing the integration of quantum channels and quantum key distribution. Numbers here represent the life-cycle of Optimus-Q.}
    \label{fig:enter-label}
\end{figure*}

\section{Methodology}

The proposed methodology encompasses several essential components designed to facilitate adaptive learning for quantum human assistant robots operating within nuclear power plants. These components are integral to ensuring that the robots can effectively learn from their environments, enhance their operational capabilities, and respond dynamically to the unique challenges posed by high-stakes settings such as nuclear facilities. By integrating these elements, we aim to create a robust framework that supports the continuous improvement and adaptability of the robots in their roles as intelligent assistants.

\subsection{Optimus-Q Responsibilities}

In this section, we provide an in-depth discussion regarding the various operational states of the Optimus-Q robot, a sophisticated system designed to enhance monitoring and safety within NPPs. Our analysis begins with the foundational assumption that the Optimus-Q robot is equipped with advanced capabilities to continuously stream sensor data in real-time, which is essential for maintaining optimal operational efficiency and safety standards in a high-stakes environment such as a nuclear facility.

The Optimus-Q robot is not merely a passive data collector; it possesses the advanced functionality to detect and predict potential contamination events within the NPP. This predictive capability is crucial, as it allows for proactive measures to be taken before contamination can pose a risk to both personnel and the environment. In this context, we elaborate on the specific types of sensor data that our system expects to receive from the Optimus-Q robot. 

\subsubsection{Sensors}

In the quest to enhance environmental monitoring in NPPs, a robot can be equipped with a variety of advanced sensors to detect contamination levels. Among these, infrared sensors are particularly important as a starting point for our experiments. These sensors are adept at detecting specific gases, such as carbon dioxide (CO$_2$), carbon monoxide (CO), and methane (CH$_4$), by measuring the absorption of infrared light at designated wavelengths. This capability allows for real-time monitoring of hazardous gas concentrations, making infrared sensors a crucial tool for early detection of potential leaks and ensuring immediate safety responses.

While other sensors, such as Geiger-Müller counters, scintillation detectors, alpha and beta particle detectors, particulate matter sensors, volatile organic compound (VOC) sensors, and gas chromatography-mass spectrometry (GC-MS), are vital for detecting various types of radiation and airborne contaminants, infrared sensors offer a non-invasive and rapid assessment of gas emissions. Starting with infrared sensors not only provides a solid foundation for understanding gas dynamics in the environment but also allows for swift identification of air quality issues, forming a critical component of a comprehensive contamination monitoring strategy in nuclear facilities.

\subsubsection{Scanning State}

One of the primary responsibilities of the Optimus-Q robot is to systematically scan the nuclear power plant while continuously monitoring for various contamination zones. This section specifically addresses this critical function. Each robot will operate within a predefined bounding box, where its task is to navigate the designated area without encroaching upon the boundaries of adjacent boxes. This approach is designed to facilitate efficient traffic management among multiple robots operating simultaneously.

The bounding boxes can take on various shapes; however, rectangular or square configurations are preferred for practical reasons. Utilizing shapes such as circles would introduce unnecessary complexity, as arranging circular bounding boxes within the nuclear power plant's layout would inevitably create gaps or voids between them. Consequently, adopting a square shape simplifies the execution of the primary scanning objective.

The specific latitudes and longitudes defining each bounding box will be programmed into every Optimus-Q robot. Additionally, each robot will initialize with a model that is downloaded from a global server, which will be further elaborated upon in subsequent sections. This model will guide the robot's operations and enhance its ability to detect and assess contamination effectively within the designated areas.

\textbf{Lawnmower Pattern Coverage.} In this section, we elaborate on the methodology by which the Optimus-Q robot will efficiently cover a square area using a systematic lawnmower pattern. We start by determining the number of strips necessary to adequately cover the width of the square area. This calculation is crucial for ensuring optimal coverage and effective navigation throughout the designated space. Let $W$ be the width of the square (in meters, calculated from latitude/longitude difference) and $w$ be the effective coverage width of the robot. The number of strips, $n$, is then given by:
\begin{equation}
n = \left\lceil \frac{W}{w} \right\rceil,
\end{equation}
where $\lceil \cdot \rceil$ denotes the ceiling function, rounding up to the nearest integer to ensure complete coverage.

Next, we determine the length of each strip. If $L$ is the length of the square (in meters, also calculated from latitude/longitude), the length of each strip, $l$, is simply:
\begin{equation}
l = L.
\end{equation}
For a perfect square, the length of each strip equals the side length of the square.

The total distance traveled by the robot, $D$, in an ideal scenario, can be calculated as:
\begin{equation}
D = n \cdot l + (n - 1) \cdot d,
\end{equation}
where $d$ represents the distance the robot travels to move from the end of one strip to the beginning of the next (the ``turn'' distance). This formula assumes direct movement between adjacent strips.

To determine the starting coordinates of each strip, we assume the bottom-left corner of the square is the origin $(0,0)$ of our local coordinate system. Let $(x_i, y_i)$ be the starting coordinates of the $i$-th strip (where $i = 0, 1, 2, \dots, n-1$). If the strips are oriented vertically, the coordinates are:
\begin{align}
x_i &= i \cdot w, \\
y_i &= 0.
\end{align}
Conversely, for horizontally oriented strips, the coordinates are:
\begin{align}
x_i &= 0, \\
y_i &= i \cdot w.
\end{align}

Finally, to calculate $W$ and $L$ from latitude and longitude coordinates, we can use a simplified version of the Haversine-style approximation for relatively small distances. Let $(lat_1, lon_1)$ and $(lat_2, lon_2)$ be the latitude and longitude of two points in decimal degrees, and let $R$ be the Earth's radius (approximately 6{,}371{,}000 meters). Define $\Delta lat = lat_2 - lat_1$ and $\Delta lon = lon_2 - lon_1$. The approximate distance between the two points (in meters) is:
\begin{equation}
\text{distance} \approx R \cdot \sqrt{\left(\Delta lat \cdot \frac{\pi}{180}\right)^2 + \left(\cos\left(lat_1 \cdot \frac{\pi}{180}\right) \cdot \Delta lon \cdot \frac{\pi}{180}\right)^2}.
\end{equation}

This approximation is most effective for shorter distances. However, when dealing with larger areas, it is essential to employ more accurate methods or libraries to ensure precise coverage and navigation. For long-range distance applications, algorithms such as A* search, Rapidly-exploring Random Trees (RRT), and Dijkstra's algorithm are particularly suitable. These algorithms can efficiently determine optimal paths and navigate complex environments, making them ideal for large-scale operations.

It is important to note that, for the purposes of our experiments and the implementation of our architecture, the speed and direction of the Optimus-Q robot are preset to specific ranges. This controlled configuration allows us to systematically evaluate the robot’s performance and adaptability under defined conditions, facilitating a more focused assessment of our methodology.

\subsubsection{Critical State}

This state is a critical component of our system, as it represents the ideal operational condition for the Optimus-Q robot when it encounters air quality contamination or detects the presence of hazardous gases. The Optimus-Q is equipped with advanced infrared sensors specifically designed to identify elevated levels of carbon dioxide (CO$_2$), carbon monoxide (CO), and excess methane (CH$_4$) emissions in the environment. The functionality of our predictive algorithms is particularly significant in this context, as they play a vital role in identifying potential contamination hotspots—referred to as ``red spots''—and conducting further tests in those areas.

The Optimus-Q continuously streams input data from its infrared sensors, allowing it to monitor air quality in real-time. This ongoing data collection enables the robot to predict concentrations of carbon dioxide, carbon monoxide, and methane gases with high accuracy. To ensure effective monitoring, threshold levels for these gases are established using collective models that are aggregated at the global server level. This aggregation process synthesizes data from multiple robots, enhancing the reliability of our predictions and ensuring that the system can respond promptly to contamination events.

\begin{figure}
\centering
\includegraphics[width=0.5\columnwidth]{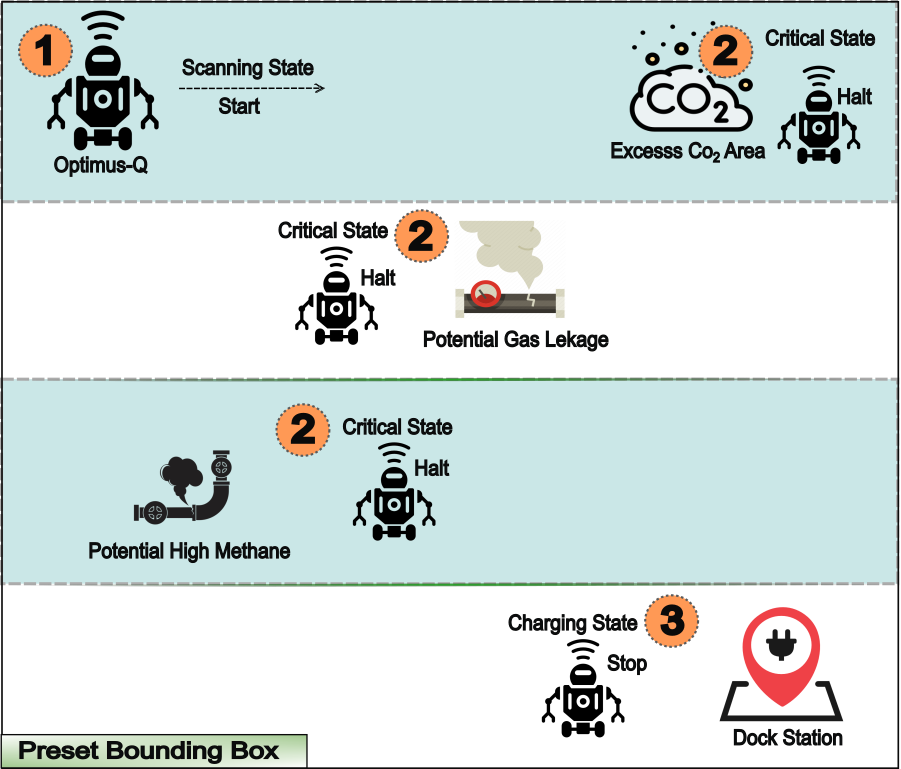}
\caption{The figure depicts the bounding box within which the Optimus-Q robot will operate to identify contamination zones.} 
\label{fig:FLvsSAFARI}
\end{figure}

Once a contamination spot is detected, the Optimus-Q robot will transmit critical information, including the latitude and longitude coordinates of the identified location, along with the specific measurements of the elevated gas levels. These measurements may include any combination of carbon dioxide, carbon monoxide, and methane, providing comprehensive data that can inform further investigation and remediation efforts. This systematic approach not only enhances the robot's ability to address air quality issues effectively but also contributes to the overall safety and operational efficiency of the nuclear power plant environment via the local server mentioned in Fig.~\ref{fig:enter-label}. Once the information is transmitted to the local server and the Optimus-Q receives an acknowledgment, it will begin to move in the specified direction and focus on the designated areas.

\subsubsection{Charging State}

The charging state is a critical component in the operation of the Optimus-Q robot, particularly for the ongoing training and updating of its predictive model. During the scanning phase, if the battery level of the Optimus-Q drops below 20\%, the robot will automatically return to its designated charging dock to recharge. For the purposes of our experiments, we have assumed that each bounding box previously discussed will contain its own charging station. However, in future iterations, there may be a centralized charging station for all Optimus-Q robots located in a single area. This centralized approach could significantly simplify maintenance procedures and streamline operations.

To reduce the complexity of our proposed system, we have implemented the approach of having preset coordinates for the charging stations stored within the Optimus-Q robot's navigation system. When the battery level reaches the critical 20\% threshold, the robot autonomously navigates to the charging dock. Upon arrival, it initiates the wireless charging process, allowing for a seamless transition into the training phase.

While charging, the Optimus-Q robot begins to train its machine learning model using historical sensor data collected during the previous operational session—defined here as the time elapsed since the last charging event. This session’s sensor data is retrieved from the robot’s internal memory, enabling the model to learn from past experiences and improve its predictive capabilities.

Once the training phase is completed and the model reaches a satisfactory performance, the robot will disconnect from the charging dock and revert to its scanning state, ready to resume its monitoring duties. However, if the robot reaches a full charge without completing the training process, it will remain docked and continue training until the update is finalized.

An additional crucial step that occurs after the model training is the interaction with the global server. The Optimus-Q robot engages in a federated learning process, allowing it to assimilate knowledge from other similar nuclear power plants in the vicinity. This collaborative learning enhances the robot's operational efficiency and adaptability, further strengthening its ability to detect and respond to contamination in its environment.

\begin{figure}
\centering
\includegraphics[width=0.5\columnwidth]{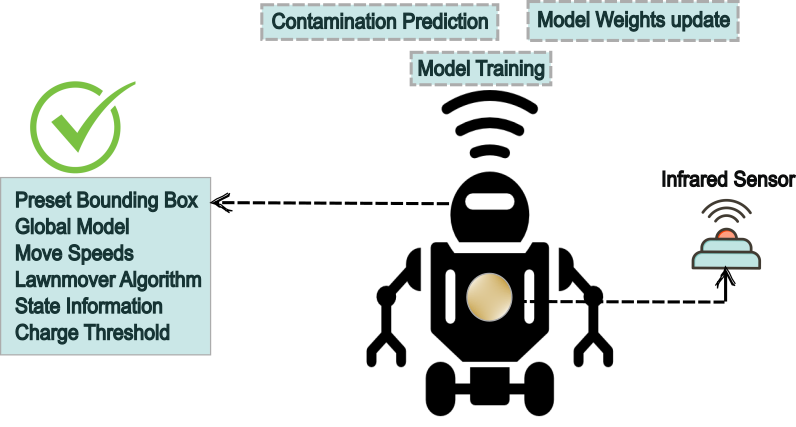}
\caption{The picture outlines the various roles and responsibilities of Optimus-Q, highlighting its preset attributes.} 
\label{fig:robot-roles}
\end{figure}

\subsection{Secure Quantum Channels}

Once the model has been successfully trained and generated using the historical session data collected by the Optimus-Q robot while it is docked for charging, we proceed with the federated learning process. This process, described later, enables the Optimus-Q to collaboratively learn from data while maintaining data privacy and security across multiple robots and facilities.

The model produced during this training phase will consist of classical bits, represented as binary values (0 or 1). However, to leverage the advantages of quantum communication, we convert this classical model into quantum states, specifically qubits. Each qubit can be assigned a spin or polarization, which represents information beyond traditional binary encoding. This transformation allows us to utilize the unique characteristics of quantum mechanics to enhance security for data transmission.

Once the model has been converted into qubits, the entire dataset will be transmitted through a secure quantum channel. At this stage, QKD will play a pivotal role in ensuring the security of the data transmission. QKD utilizes the principles of quantum mechanics to generate and distribute cryptographic keys that can be used for encrypting the data being sent. One of the most widely used QKD protocols is the BB84 protocol, which operates on the principle of superposition and the no-cloning theorem.

The process of implementing QKD within the Optimus-Q system involves several critical steps. First, during the \textbf{key generation} phase, the Optimus-Q robot creates a secret key using a series of qubits that are transmitted over a quantum channel. Each qubit can exist in a superposition of states, mathematically represented as:
\[
|\psi\rangle = \alpha |0\rangle + \beta |1\rangle,
\]
where $|\psi\rangle$ is the state of the qubit, $|0\rangle$ and $|1\rangle$ are the basis states, and $\alpha$ and $\beta$ are complex coefficients that satisfy the normalization condition $|\alpha|^2 + |\beta|^2 = 1$.

Following this, in the \textbf{key distribution} stage, the generated key is transmitted to the global server via the quantum channel. Importantly, if an eavesdropper attempts to intercept the qubits, any measurement conducted on them will disturb their state, thus alerting the communicating parties to the presence of the eavesdropper. The disturbance can be quantified using the quantum bit error rate (QBER), defined as:
\[
\text{QBER} = \frac{E}{N},
\]
where $E$ is the number of erroneous bits detected, and $N$ is the total number of bits transmitted. A higher QBER indicates a greater likelihood of interception.

Once the key is securely distributed, the next step is \textbf{data encryption}. The secret key is utilized to encrypt the classical bits of the model, ensuring that the data remains confidential during transmission. This encryption process can be mathematically represented by the one-time pad encryption method, where the encrypted message $E$ is expressed as:
\[
E = M \oplus K,
\]
in which $M$ represents the original message (the classical model bits), $K$ is the secret key generated via QKD, and $\oplus$ denotes the bitwise XOR operation. The security of this encryption method relies on the key $K$ being truly random and at least as long as the message $M$.

Finally, during the \textbf{transmission} phase, the encrypted data $E$ is sent through the quantum-secured classical channel, guaranteeing that the integrity and confidentiality of the data are upheld throughout the transfer process. The effectiveness of the QKD protocol can be further illustrated by the following inequality, which ensures the security of the shared key:
\[
I(A:E) \leq S,
\]
where $I(A:E)$ is the mutual information between Alice and Eve (the eavesdropper), and $S$ is the secrecy capacity of the channel. This inequality demonstrates that the information available to the eavesdropper must be minimized to maintain the confidentiality of the key.

Upon reaching the global server, we decrypt the transmitted data using the shared key. The ability to share model information with quantum-secured keys not only enhances the security of data handling but also opens new avenues for improving the predictive capabilities of the Optimus-Q robot through collective learning across various nuclear power plants.

The integration of QKD in this context is particularly effective because it provides a level of security that is unattainable with classical cryptographic methods. The inherent properties of quantum mechanics ensure that any attempt to eavesdrop will be detectable, thus safeguarding sensitive data throughout its transmission. This innovative approach to model training and data transmission exemplifies the potential of integrating quantum technologies with robotic systems, ultimately leading to more robust and intelligent monitoring solutions in high-stakes environments such as nuclear facilities.

\subsubsection{Technical Drawbacks}

One of the drawbacks we have identified through our experiments is that while integrating QKD into a communication system, such as that utilized by the Optimus-Q robot, offers considerable security benefits, it also introduces various latency-related challenges. One key issue is the time-consuming nature of key generation, as QKD relies on the transmission and measurement of qubits, which can delay the overall process, especially over long distances or in the presence of noise and signal loss in the quantum channel.

Additionally, after generating a key, QKD often necessitates further steps such as error correction and privacy amplification to ensure key reliability and security, which involve additional computations and communications that contribute to latency. The performance of QKD is highly dependent on the quality of the quantum channel, and factors such as attenuation and environmental interference can exacerbate delays. Furthermore, establishing a QKD connection requires an initial setup phase to configure the quantum channel and agree on parameters for key generation, adding significant latency, particularly in dynamic environments where Optimus-Q may frequently need to establish connections.

QKD systems also tend to be resource-intensive, requiring sophisticated hardware and protocols that may not be optimized for speed; this complexity can slow data exchange compared to classical key distribution methods, which are generally faster due to simpler infrastructures. Finally, in scenarios where real-time data transmission is critical—such as in monitoring systems like the Optimus-Q robot—any added latency from the QKD process could impact the system's responsiveness, hindering the robot's ability to react quickly to contamination events or other critical situations. Therefore, while QKD offers unparalleled security advantages, its integration necessitates careful consideration of optimizing the QKD process, improving quantum channel quality, and ensuring that the overall system architecture can accommodate the inherent delays associated with quantum key generation and distribution.

\subsection{Local Server}

The primary objective of the local server is to facilitate technicians' access to monitoring data and deliver near real-time updates regarding environmental conditions within the nuclear facility. This capability is particularly crucial when it comes to promptly addressing any instances of contamination detected by the Optimus-Q robot.

\subsubsection{Monitoring Systems}

The local server receives data whenever the Optimus-Q robot identifies signs of environmental contamination, leveraging its advanced sensing technology. Specifically, the robot is equipped with infrared sensors that are adept at detecting elevated levels of gases such as carbon dioxide (CO$_2$), carbon monoxide (CO), and methane (CH$_4$). These sensors play a vital role in ensuring that any hazardous changes in air quality are rapidly recognized and communicated to the local server.

When contamination is detected, the transmission of data to the local server will encompass several key pieces of information. This includes the geographical coordinates, specifically the latitude and longitude, which are essential for accurately pinpointing the location of the contamination. In addition to the location data, the transmission will contain precise measurements of the gas levels for carbon monoxide, carbon dioxide, and methane. These readings are critical for assessing the severity of the contamination and determining the appropriate response measures.

To enhance the utility of the monitoring data, each transmission will also include a timestamp, providing context regarding when the contamination was detected. This temporal information is invaluable for tracking the progression of contamination events and for conducting thorough analyses of air quality over time. By consolidating this data, the local server ensures that technicians are equipped with timely and relevant information, enabling them to make informed decisions and take immediate action in response to any environmental hazards identified by the Optimus-Q robot.

\subsubsection{Secure Data Transmission}

The ChaCha encryption algorithm is a modern and efficient stream cipher that can be utilized for securing data transmitted from the Optimus-Q robot to the local server. This algorithm is particularly well-suited for environments requiring high-speed encryption and decryption, as it maintains both security and performance. When the Optimus-Q robot detects environmental contamination, the data collected—including latitude, longitude, gas measurements (carbon dioxide, carbon monoxide, and methane levels), and timestamps—needs to be securely transmitted to the local server. Before transmission, this sensitive information is encrypted using the ChaCha algorithm, ensuring that the data remains confidential and protected against potential eavesdropping or tampering.

The encryption process involves the generation of a unique key and nonce for each session, which enhances security by preventing replay attacks. Once the data is encrypted, it is transmitted over the communication channel to the local server, where it can be decrypted and analyzed by technicians. By employing the ChaCha algorithm, the system ensures that real-time monitoring data is transmitted securely, allowing for immediate and informed responses to any environmental hazards detected by the Optimus-Q robot.

\subsection{Global Server}

The goal of the global server is to perform the federated aggregation received from different Optimus-Q robots across various nuclear power plants (NPPs). We have opted for FedAvg at the global server level to perform our operations.

\subsubsection{Federated Learning}

Federated learning is a decentralized machine learning paradigm that enables multiple devices to collaboratively learn a shared model while keeping their training data local. This approach is particularly relevant in environments where data privacy, security, and compliance are critical concerns, such as in nuclear power plants. By allowing the Optimus-Q robot and other similar devices to learn from their experiences without transferring sensitive data to a central server, federated learning enhances data privacy and reduces the risk of exposing confidential information.

In our system, once the Optimus-Q robot successfully completes its training process and generates its local model while connected to the charging dock, the subsequent step involves transmitting the acquired local model weights to the global server. This transmission is facilitated through a secure channel whose keys are established using QKD to ensure that the data is transmitted safely across the network. By utilizing QKD to provision encryption keys, we protect the model weights from potential interception and unauthorized access, thereby safeguarding the integrity and confidentiality of the sensitive information being exchanged.

Let $w_t$ represent the global model weights at iteration $t$, and let $w_i^t$ denote the local model weights of the $i$-th Optimus-Q robot after performing local training on its dataset for $E$ epochs. The FedAvg algorithm proceeds as follows:

\begin{enumerate}[leftmargin=1.2em]
    \item Each robot $i$ initializes its model with the current global weights $w_t$.
    \item Each robot conducts local training using its own data for a defined number of epochs $E$, resulting in updated weights $w_i^t$. A simple local update can be written as
    \begin{equation}
        w_i^t = w_t - \eta \nabla F_i(w_t),
    \end{equation}
    where $\eta$ is the learning rate, and $\nabla F_i(w_t)$ is the gradient of the local loss function for robot $i$ evaluated at $w_t$.
    \item After completing the local training, the robots transmit their updated weights back to the global server. The server aggregates these weights using a weighted average based on the number of data samples available to each robot $n_i$:
    \begin{equation}
        w_{t+1} = \sum_{i=1}^{N} \frac{n_i}{\sum_{j=1}^{N} n_j} w_i^t,
    \end{equation}
    where $N$ denotes the total number of participating robots.
\end{enumerate}

The aggregation process is critical, as it ensures that the global model reflects the insights and learning gleaned from the diverse operational environments of all participating robots. Once the aggregation is complete, the resultant global model weights are transmitted back to the client nodes, which in this case are the Optimus-Q robots. This updated global model, enriched with knowledge from multiple clients, allows the robots to enhance their understanding of environmental conditions and improve their predictive capabilities regarding contamination detection. By leveraging insights from a broader dataset, the Optimus-Q robots can better identify and anticipate areas of potential contamination, leading to more effective monitoring and response strategies in NPPs. This collaborative learning framework not only optimizes the performance of individual robots but also contributes to a more robust and intelligent overall monitoring system.

\subsubsection{Convergence Considerations}

The convergence of federated learning, particularly when utilizing the FedAvg algorithm, is essential for determining the effectiveness and efficiency of the model training process, especially in the context of the Optimus-Q robot. In our implementation, multiple Optimus-Q robots operate in various locations, each collecting environmental data, including measurements of CO$_2$, CO, and CH$_4$ levels. The goal is to collaboratively learn a global model that reflects the collective knowledge gained from diverse operational environments while ensuring that individual data remains secure and private.

Under common assumptions such as smoothness of the local loss functions and bounded variance of stochastic gradients, FedAvg can be shown to converge toward a critical point of the global loss function $F(w)$, defined as a weighted sum of local losses:
\begin{equation}
F(w) = \sum_{i=1}^{N} \frac{n_i}{\sum_{j=1}^{N} n_j} F_i(w),
\end{equation}
where $F_i(w)$ is the local loss for robot $i$. While exact convergence rates depend on step sizes, data heterogeneity, and the number of local steps, practical deployments can empirically verify convergence through monitoring training and validation losses across communication rounds.

\begin{figure}
\centering
\includegraphics[width=0.7\columnwidth]{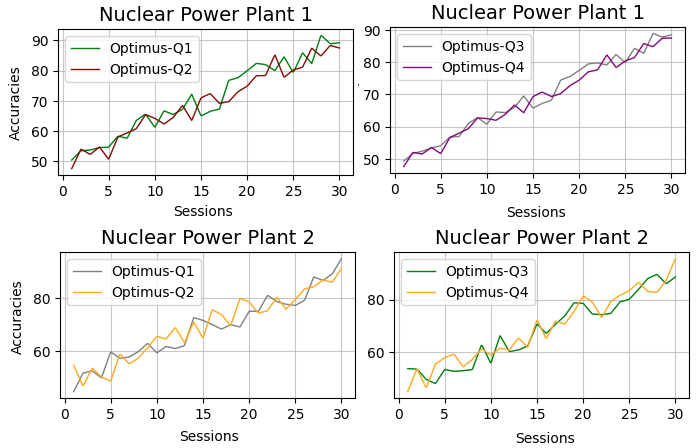}
\caption{The picture outlines the various roles and responsibilities of Optimus-Q, highlighting its preset attributes.} 
\label{fig:robot-roles}
\end{figure}

%%%%%%%%%%%%%%%%%%%%%%%%%%%%%%%%%%%%%%%%%%%%%%%%%%%%%%%%%%%%%%%%%%%%%%%%%%%%%%%%
\section{Experiments}

This section elucidates the experimental procedures employed to validate the proposed methodology for the Optimus-Q robot, focusing on its interactions with the global server, the data collection process, contamination detection, and secure data transmission.

The experiments were designed to evaluate the performance of the Optimus-Q robot in a simulated nuclear power plant environment. The primary goal was to assess its ability to autonomously receive an updated operational model from the global server, perform real-time data collection using infrared sensors, detect contamination events, and securely transmit crucial information back to the server. We considered two NPP setups, with each plant featuring four distinct Optimus-Q robots. The experiments were conducted with these eight Optimus-Q robots over a total of 30 epochs (or sessions), where one session corresponds to a full charge-discharge cycle.

\begin{table*}[ht]
    \centering
    \setlength{\tabcolsep}{1pt}
    \begin{tabular}{|l|lllll|lllll|}
    \hline
    \textbf{} & \multicolumn{5}{c|}{\textbf{Nuclear Power Plant 1}} & \multicolumn{5}{c|}{\textbf{Nuclear Power Plant 2}} \\ 
    \hline
    \textbf{Run Name} & \textbf{Accuracy} & \textbf{F1 Score} & \textbf{Precision} & \textbf{Recall} & \textbf{ROC AUC} & \textbf{Accuracy} & \textbf{F1 Score} & \textbf{Precision} & \textbf{Recall} & \textbf{ROC AUC} \\ 
    \hline
    Session 1  & 0.586667 & 0.490355 & 0.452679 & 0.586667 & 0.991834 & 0.466667 & 0.345518 & 0.297651 & 0.466667 & 0.986534 \\ 
    \hline
    Session 5  & 0.600000 & 0.519155 & 0.488471 & 0.600000 & 0.999101 & 0.500000 & 0.390684 & 0.351193 & 0.500000 & 0.988069 \\ 
    \hline
    Session 10 & 0.613333 & 0.524141 & 0.489656 & 0.613333 & 0.993721 & 0.553333 & 0.453966 & 0.420563 & 0.553333 & 0.998413 \\ 
    \hline
    Session 15 & 0.753333 & 0.675188 & 0.635296 & 0.753333 & 0.999418 & 0.653333 & 0.571602 & 0.538000 & 0.653333 & 0.997703 \\ 
    \hline
    Session 20 & 0.786667 & 0.718362 & 0.680571 & 0.786667 & 0.999295 & 0.766667 & 0.695252 & 0.660074 & 0.766667 & 0.999788 \\ 
    \hline
    Session 25 & 0.867333 & 0.862952 & 0.848691 & 0.880930 & 0.999894 & 0.873333 & 0.829886 & 0.805349 & 0.873333 & 0.999788 \\ 
    \hline
    Session 30 & 0.897800 & 0.893444 & 0.879778 & 0.920000 & 0.999788 & 0.861012 & 0.857952 & 0.838111 & 0.893333 & 0.999894 \\ 
    \hline
    \end{tabular}
    \caption{Comparison metrics between Nuclear Power Plant 1 and Nuclear Power Plant 2 based on performance across thirty sessions. Each session is defined as one full charge-to-charge cycle of an Optimus-Q robot.}
    \label{tab:ml-comparison-metrics}
\end{table*}

\subsection{Model Retrieval from the Global Server}

Upon being powered on, the Optimus-Q robot establishes a connection to the global server with the primary objective of requesting the most up-to-date predictive model. This model is of paramount importance, as it encompasses the necessary algorithms and parameters that enable the robot to perform its tasks with optimal effectiveness. Among the initial parameters that the Optimus-Q robot comes equipped with are several key components: the bounding box, which defines the operational area; the lawnmower pattern algorithm, which dictates the robot's movement strategy; movement speeds that determine how quickly the robot can traverse its environment; a low charging threshold to manage its power efficiency; and various security algorithms designed to ensure safe operation within its designated area.

\subsection{Data Stream and Sensor Integration}

With the operational model firmly established, the Optimus-Q robot is fully equipped to initiate its primary function: environmental monitoring within the nuclear facility. This critical task encompasses a comprehensive data collection process, which is essential for ensuring the safety and integrity of the environment in which it operates.

The robot continuously gathers data from its onboard infrared sensors. These sensors measure concentrations of various gases that can indicate hazardous conditions—specifically, CO$_2$, CO, and CH$_4$. This real-time data acquisition is vital for maintaining stringent safety standards within the nuclear facility, where even minor fluctuations in gas concentrations could signal potential risks.

Once the data is collected, it undergoes a processing phase locally within the robot's system. This processing allows for the assessment of air quality in the immediate environment. Additionally, the robot employs predictive algorithms to anticipate potential contamination events. Such proactive predictions are invaluable, as they enable technicians to respond swiftly and effectively, reaching the contaminated area in a timely manner to mitigate adverse effects.

In scenarios where the robot detects any signs of contamination—be it through abnormal gas concentrations or other environmental indicators—it is programmed to take immediate action. The robot will either halt its movements to prevent further exposure or shift into a critical state, serving as an urgent alert mechanism that signifies that immediate attention is required. By implementing this responsive strategy, Optimus-Q enhances its operational effectiveness and plays a vital role in safeguarding the health and safety of personnel working within the facility.

\subsection{Contamination Detection Process}

When the Optimus-Q robot detects that contamination levels exceed predefined safety thresholds, it promptly gathers and transmits vital data, including its geographic coordinates—specifically, the latitude and longitude of the contaminated area—alongside the excess levels of harmful gas concentrations detected by its infrared sensors. These measurements encompass key pollutants such as CO$_2$, CO, and CH$_4$. To ensure the secure transmission of this sensitive data from the Optimus-Q robot to the local server, the ChaCha secure encryption algorithm is employed, providing a robust layer of security against potential unauthorized access or data tampering during the communication process. This meticulous approach underscores the robot's commitment to safety and efficiency in its operational duties.

\subsection{Secure Data Transmission to the Local Server}

The process commences with the critical phase of data preparation, during which the robot meticulously gathers essential information required for subsequent transmission. This information includes various parameters such as contamination levels, geographical coordinates that pinpoint the exact location of the contamination, and timestamps that specify when the contamination was detected. These elements are fundamental for ensuring accurate tracking and response to contamination events.

Once the data collection is complete, the robot encrypts the gathered information using a secure key via the ChaCha algorithm. The encoded data is then transmitted to a local server through conventional communication channels, which may include Wi-Fi or cellular networks. The use of encryption serves as a robust protective measure, ensuring that even if the data is intercepted during transmission, it remains unreadable without the corresponding decryption key.

Upon successful receipt of the transmitted data, the local server processes and logs the information. This step involves analyzing the data for further insights and maintaining a record for future reference. Following the processing, the server sends an acknowledgment back to the robot, confirming that the data has been received successfully. This acknowledgment is crucial as it enables the robot to continue its operations seamlessly, without interruption.

\subsection{Secure Data Transmission to the Global Server}

The secure data transmission process to the global server is a critical component of the Optimus-Q robot's operational framework, particularly in the context of nuclear power plants where data integrity and confidentiality are paramount. Transmission begins after the Optimus-Q robot has completed its local training and generated model weights based on its environmental data collection.

To initiate the transmission, the Optimus-Q robot connects to its charging dock, where it prepares to send the acquired local model weights to the global server. As discussed earlier, keys generated via QKD can be used to encrypt these weights, ensuring that the data is transmitted safely across the network. Libraries such as \texttt{Qiskit} or \texttt{PyQuil} can be used in a simulation environment to prototype quantum key establishment, though deployment would depend on dedicated quantum hardware links.

Once the encrypted local model weights are transmitted, the global server receives the data and initiates the FedAvg aggregation process. After aggregation, the updated global model weights are sent back to the Optimus-Q robots, again protected with keys provisioned via QKD where available.

\subsection{Performance Metrics}

We evaluated performance metrics through two server levels: the local server and the global server. At the local server level, our focus was on comprehensively understanding the contamination areas and assessing the accuracy of the data associated with them. This involved analyzing the information collected by the Optimus-Q robots to determine the precision with which these areas were identified and characterized.

At the global server level, we adopted a broader perspective to measure accuracy through the convergence of machine learning models sourced from various Optimus-Q robots. This approach allowed us to aggregate and synthesize insights from multiple data streams, thereby enhancing the overall reliability of our assessments. By comparing and integrating the outputs from these diverse models, we achieved a more holistic understanding of contamination dynamics across the entire operational landscape. This dual-level evaluation not only facilitated a nuanced analysis of local contamination issues but also contributed to the overarching goal of improving predictive accuracy and operational effectiveness at a global scale.

%%%%%%%%%%%%%%%%%%%%%%%%%%%%%%%%%%%%%%%%%%%%%%%%%%%%%%%%%%%%%%%%%%%%%%%%%%%%%%%%
\section{Conclusion}

This paper presented the Optimus-Q robot, an advanced system designed to enhance safety and monitoring in nuclear power plants through adaptive learning and secure quantum communication. Our methodology integrates real-time environmental data processing and predictive analytics, enabling the robot to autonomously detect and respond to contamination events effectively.

Experimental results across two simulated nuclear power plant setups demonstrate that Optimus-Q significantly improves contamination monitoring accuracy and efficiency. By leveraging advanced infrared sensors, the robot can identify hazardous gases like CO$_2$, CO, and CH$_4$ in real-time, facilitating timely safety measures. Federated learning enhances the robot's predictive capabilities without compromising data privacy, allowing it to learn from diverse operational environments. Additionally, the integration of QKD strengthens data security during model exchange, addressing critical information security concerns.

Overall, the Optimus-Q framework illustrates the potential of combining robotics, federated learning, and quantum-secured communication to build intelligent, resilient monitoring systems for high-stakes environments. Future work includes deploying Optimus-Q in physical testbeds, optimizing QKD latency, and extending the framework to multi-modal sensing and decision-support systems.

%%%%%%%%%%%%%%%%%%%%%%%%%%%%%%%%%%%%%%%%%%%%%%%%%%%%%%%%%%%%%%%%%%%%%%%%%%%%%%%%

\bibliographystyle{unsrtnat} % or plainnat, abbrvnat, etc.
\bibliography{references}

\end{document}